\documentclass[letterpaper]{article} 
\usepackage{aaai25}  
\usepackage{times}  
\usepackage{helvet}  
\usepackage{courier}  
\usepackage[hyphens]{url}  
\usepackage{graphicx} 
\usepackage{natbib}  
\usepackage{caption} 
\usepackage{algorithm}
\usepackage{algorithmic}

\usepackage{amsmath}
\usepackage{amsfonts}
\usepackage{subcaption}
\usepackage{pifont}
\usepackage{booktabs}
\usepackage{multirow}
\usepackage{colortbl}
\usepackage{xcolor} 
\newcommand{\cmark}{\ding{52}}%
\newcommand{\xmark}{\ding{56}}%

\usepackage{newfloat}
\usepackage{listings}
\DeclareCaptionStyle{ruled}{labelfont=normalfont,labelsep=colon,strut=off} 
\floatstyle{ruled}
\newfloat{listing}{tb}{lst}{}
\floatname{listing}{Listing}
\title{Exploring Temporal Event Cues for Dense Video Captioning in Cyclic Co-learning}
\author{
    Zhuyang Xie\textsuperscript{\rm 1}, 
    Yan Yang\textsuperscript{\rm 1,2}\thanks{Corresponding author.}, 
    Yankai Yu\textsuperscript{\rm 1},
    Jie Wang\textsuperscript{\rm 1},
    Yongquan Jiang\textsuperscript{\rm 1,2},
    Xiao Wu\textsuperscript{\rm 1,2}\\
}
\affiliations{
    \textsuperscript{\rm 1}School of Computing and Artificial Intelligence, Southwest Jiaotong University, Chengdu, China \\
    \textsuperscript{\rm 2}Engineering Research Center of Sustainable Urban Intelligent Transportation, Ministry of Education, China \\
    \{zyxie, yyk\}@my.swjtu.edu.cn, 
    \{yyang, jackwang, yqjiang, wuxiaohk\}@swjtu.edu.cn
}

\usepackage{bibentry}

\begin{document}

\maketitle

\begin{abstract}
Dense video captioning aims to detect and describe all events in untrimmed videos. This paper presents a dense video captioning network called Multi-Concept Cyclic Learning (MCCL), which aims to: (1) detect multiple concepts at the frame level and leverage these concepts to provide temporal event cues; and (2) establish cyclic co-learning between the generator and the localizer within the captioning network to promote semantic perception and event localization. Specifically, we perform weakly supervised concept detection for each frame, and the detected concept embeddings are integrated into the video features to provide event cues. Additionally, video-level concept contrastive learning is introduced to produce more discriminative concept embeddings. In the captioning network, we propose a cyclic co-learning strategy where the generator guides the localizer for event localization through semantic matching, while the localizer enhances the generator’s event semantic perception through location matching, making semantic perception and event localization mutually beneficial. MCCL achieves state-of-the-art performance on the ActivityNet Captions and YouCook2 datasets. Extensive experiments demonstrate its effectiveness and interpretability.
\end{abstract}

%

\section{Introduction}
In recent years, with the rapid growth of video data, automatically generating high-quality video captions has emerged as a significant research focus in the fields of computer vision and natural language processing. Dense video captioning~\cite{krishna2017dense,shen2017weakly,mun2019streamlined,suin2020efficient,deng2021sketch}, a critical task within this domain, requires systems not only to accurately identify multiple events but also to generate natural language descriptions for each event. The primary challenge is to simultaneously handle video content understanding, temporal event localization, and event captioning.

\begin{figure}[t]
	\centering
	\includegraphics[width=0.48\textwidth]{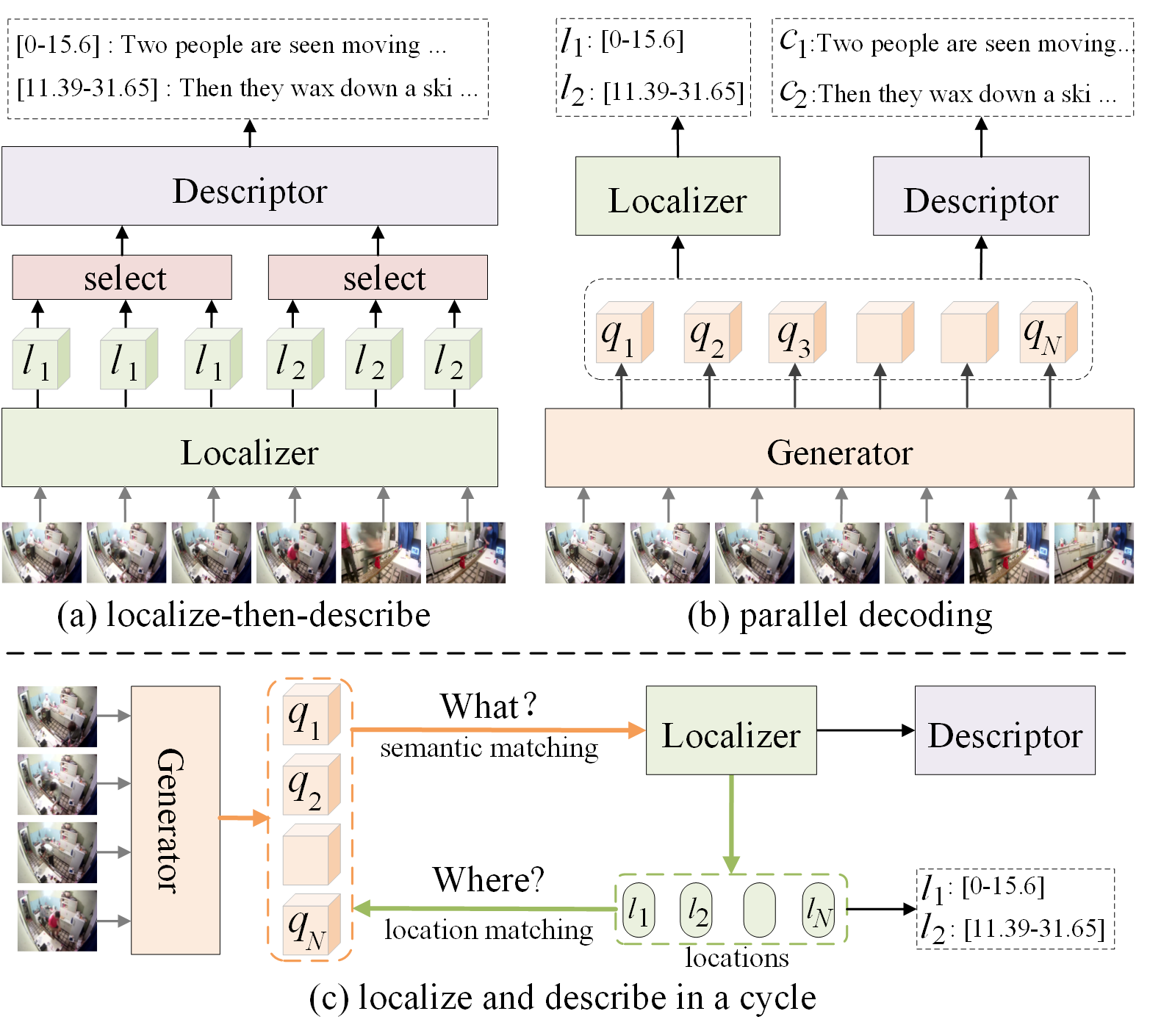}
	\caption{(a) The two-stage approach predicts multiple event candidates and eliminates redundancy using non-maximum suppression. (b) Learnable event queries are embedded in the generator (visual transformer), with two prediction heads for localization and captioning. (c) MCCL uses cyclic co-learning, where the generator, localizer, and descriptor enhance performance collaboratively by leveraging the mutual benefits of captioning and localization.}
	\label{fig:motivation}
\end{figure}

Despite recent advancements in dense video captioning, existing methods still exhibit limitations when handling complex video scenarios. Early methods typically adopt unimodal information (e.g., visual features) as input~\cite{krishna2017dense,tu2021r,zhou2018end,tu2023self} for image semantic understanding. Additionally, current approaches face challenges in feature representation and cross-modal alignment, which can lead to inconsistencies between the generated captions and video content. Inspired by recent progress in vision and language learning~\cite{radford2021learning,luo2022clip4clip,xie2023ra,deng2023prompt}, we enhance video semantic comprehension by retrieving text information that is highly relevant to the video content, thereby improving both event localization and caption quality.

Event localization and caption generation are commonly integrated into a unified framework. Mainstream methods can be divided into two categories: two-stage method~\cite{krishna2017dense,duan2018weakly,chen2021towards,iashin2020multi} and parallel decoding method~\cite{wang2021end,kim2024you}. As illustrated in Fig.\ref{fig:motivation}(a), the two-stage method initially employs a localizer to produce a large set of event proposals, which are then filtered using non-maximum suppression (NMS) to select the most accurate ones. Subsequently, a descriptor produces captions based on the visual context of these proposals. This approach heavily depends on anchor design, resulting in substantial computational overhead and limiting end-to-end training capabilities. More recently, a parallel end-to-end method PDVC~\cite{wang2021end}, shown in Fig.\ref{fig:motivation}(b), decouples dense video captioning into two parallel tasks that operate on intermediate features (event queries). However, we argue that there is a lack of an effective interaction mechanism between event semantic perception and localization. Fig.\ref{fig:motivation}(c) illustrates our cyclic co-learning mechanism, in which the generator perceives potential events (What ?) through semantic matching and guides the localizer in event localization. The localizer then feeds back the location matching to the generator (Where ?), enabling it to better perceive events. This interactive mechanism makes semantic perception and event localization mutually beneficial.

In this paper, we propose a dense video captioning network called Multi-Concept Cyclic Learning (MCCL). The network first enhances video features through video-text retrieval. It then detects concepts at the frame level to provide temporal event cues that enhance event localization and captioning. Finally, cyclic co-learning is established between the generator and localizer, enabling mutual benefits in semantic perception and event localization. The key contributions are as follows:
\begin{itemize}
	\setlength{\itemsep}{0pt}
	\item We build a cyclic mechanism between the generator and localizer to promote co-learning of semantic perception and event localization.
	\item To explore temporal event cues, multiple concepts are detected at the frame level in a weakly supervised manner, which improves both event localization and caption quality.
	\item Experiments conducted on ActivityNet Captions and YouCook2 demonstrate that the proposed method achieves state-of-the-art performance.
\end{itemize}

\section{Related Work}
\label{sec:related}

\subsubsection{Dense Video Captioning.}
\label{subsec:infounits}
Dense video captioning involves two main tasks: event localization and captioning. Early methods typically adopt a two-stage framework~\cite{krishna2017dense}, where event localization and captioning are performed separately. Some approaches aim to improve event representation to produce more informative captions~\cite{wang2018bidirectional,wang2020event,ryu2021semantic}. To achieve more robust and context-aware representations, recent studies incorporate multimodal inputs~\cite{rahman2019watch,chang2022event,aafaq2022dense}. However, these two-stage methods often fail to optimize event localization and captioning together, leading to a lack of interaction between the two tasks. To address this, PDVC~\cite{wang2021end} redefines dense video captioning as a set prediction problem, simultaneously optimizing both tasks on shared intermediate features. Some studies enhance the captioning network by using cycle consistency~\cite{Kim2021viewpoint,yue2024multi} to match generated captions with image features. In this study, we emphasize the guidance and feedback between semantic perception and event localization. Motivated by~\cite{tian2021cyclic}, we introduce a cyclic co-learning mechanism that strengthens the interaction between localization and captioning, enabling more accurate event localization and high-quality video captions.

\subsubsection{Retrieval-Enhanced Captioning.} 
In recent years, retrieval-based methods have emerged as a promising approach to enhance video captioning by improving the informativeness of generated captions~\cite{zhang2021open, yang2023concept, chen2023retrieval, jing2023memory,kim2024you}. By aligning video content with semantically relevant textual information, these methods enhance contextual understanding, addressing challenges such as insufficient context and ambiguous descriptions. This alignment enables the generation of more accurate and informative captions. In this paper, we retrieve text associated with video frames to provide additional semantic information. The retrieved sentences are then integrated with video features, facilitating more comprehensive video understanding.

\begin{figure*}[t]
	\centering
	\includegraphics[width=\textwidth]{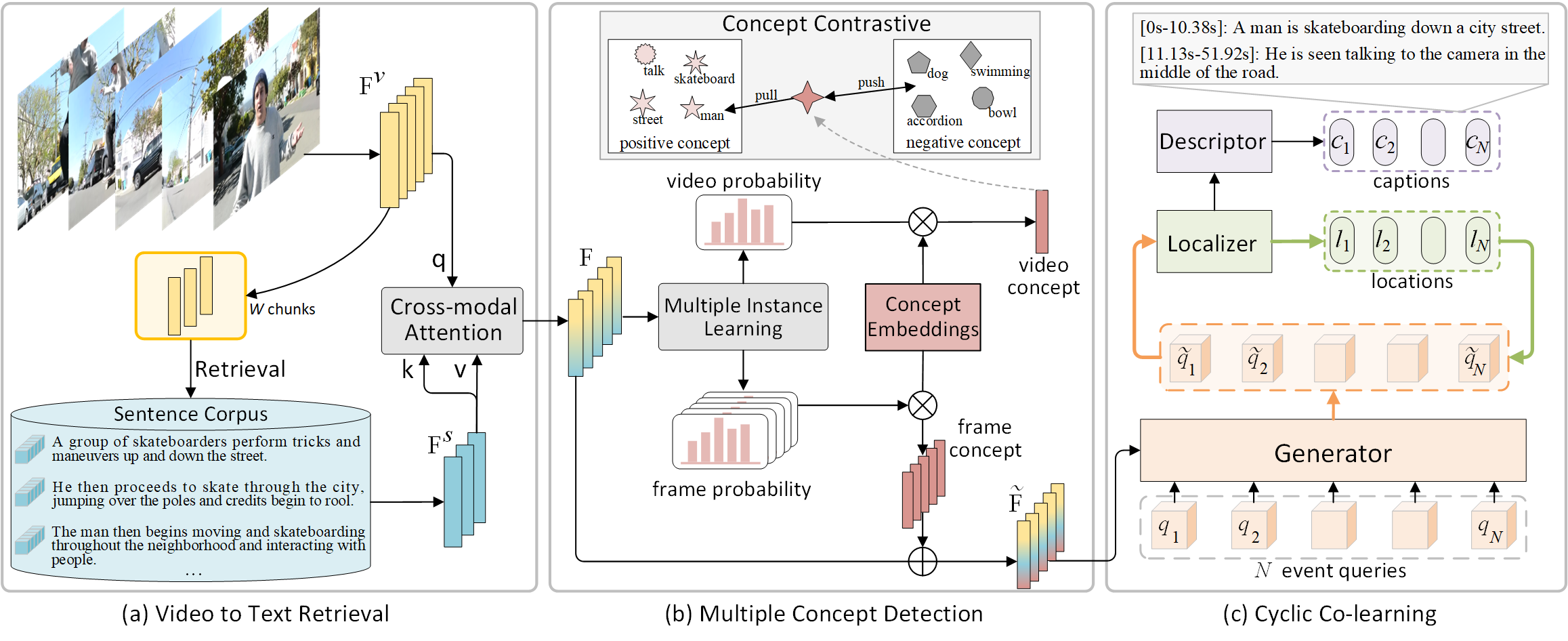}
	\caption{Overview of the framework. (a) A pretrained image encoder extracts video features and performs cross-modal retrieval to obtain sentence features. (b) Video-level and frame-level concepts are detected via multiple instance learning. (c) The features are fed into the generator to update event queries. The localizer predicts locations for each query and selects the optimal ones, while the descriptor produces captions based on these optimal queries. The generator and localizer co-learn in a cycle.}
	\label{fig:model}
\end{figure*}

\subsubsection{Concept Detection for Video Captioning.}
\label{subsec:supervision}
Concept detection plays a pivotal role in enhancing video captioning by providing a deeper understanding of video content. Recent studies have demonstrated its potential to improve both the quality and informativeness of captions~\cite{gao2020fused,yang2023concept,wu2023concept,lu2024set}. These approaches leverage video-level concepts to generate more detailed and contextually enriched captions. In this work, we apply frame-level concept detection in a weakly supervised manner, enabling the model to identify concepts such as objects, actions, and scenes within individual frames. By incorporating these concepts, our model captures fine-grained details and temporal relationships among video elements, leading to more accurate and contextually relevant captions.

\section{Methodology}
\label{sec:method}

As shown in Fig.\ref{fig:model}, the proposed MCCL consists of three components: video-to-text retrieval, multiple concept detection, and cyclic co-learning. The details are described in the following subsections.

\subsection{Video-to-Text Retrieval}
To collect semantic information as prior knowledge, a sentence corpus is constructed from in-domain training sets. By using a pretrained CLIP text encoder, sentence features are extracted and stored in the corpus. The corpus is denoted as $U=\{u_j\}^{M}_{j=1}$, where $u_j \in \mathbb{R}^{1 \times d}$ represents the $j$-th sentence feature, and $M$ is the total number of sentences.

Given an input video $V$ with $T$ frames, a pretrained CLIP image encoder extracts frame-level visual features $F^{v} = \{f^v_t\}_{t=1}^{T}$, where $f^v_t \in \mathbb{R}^{1 \times d}$. To reduce the computational cost during video-to-text retrieval, these features are evenly divided into $W$ temporal chunks, and the features within each chunk are averaged to produce chunk-level visual features $\{s_{i}\}_{i=1}^{W}$, where $s_{i} \in \mathbb{R}^{1 \times d}$. These chunk features serve as input queries for text retrieval:
\begin{equation}
\mathrm{sim}(s_{i}, u_j) = \frac{s_{i} \cdot u_j}{ ||s_{i}|| \cdot ||u_j||}, \label{eq.1}
\end{equation}
where $\mathrm{sim}(\cdot, \cdot)$ indicates the cosine similarity. For each chunk $s_{i}$, the top $N_K$ sentences are retrieved from corpus $U$ with respect to their similarity scores. These $N_K$ sentence features are subsequently mean pooled to obtain the semantic feature $f^{s}_{i} \in \mathbb{R}^{1 \times d}$. All chunk-level semantic features are collectively denoted as $F^{s} = \{f^{s}_{i}\}^{W}_{i=1}$.

To effectively integrate semantic information with visual features, cross-modal attention is applied for dynamic fusion:
\begin{small}
	\begin{equation}
	F = \mathrm{softmax}\left( \frac{F^{v} (F^{s})^{\top}}{\sqrt{d}} \right)F^{s}, \label{eq.2}
	\end{equation}
\end{small}
where $F \in \mathbb{R}^{T \times d}$ represents the aggregated video features.

\subsection{Multiple Concept Detection}
Concept detection can provide valuable guidance for localization~\cite{chen2021towards} and caption generation~\cite{yang2023concept}. Frame-level concept detection is introduced to explore temporal event cues. Specifically, the top $N_C$ most frequent words (nouns, adjectives, and verbs) from training captions are selected as concepts $E=\{e_1, e_2, \dots, e_{N_C}\}$. Then, for each video, a video-level concept label $Y^C \in \{0,1\}^{N_C}$ is constructed, indicating the presence of each concept in the ground-truth captions. Specifically, $Y^C$ is a multi-hot label where $Y^C_i=1$ if concept $e_i$ is present and $Y^C_i=0$ otherwise.

\subsubsection{Multiple Instance Learning}
Multiple concept detection takes the video features $F=\{f_t\}^{T}_{t=1}$ as input and predicts both video-level and frame-level concepts. For frame-level concept detection, a shared fully connected layer followed by a sigmoid function predicts the probability:
\begin{equation}
p_t = \sigma(FC(f_t)), \label{eq.3}
\end{equation}
where $p_t \in \mathbb{R}^{1 \times N_C}$ represents the concept probability at the frame $t$, $FC(\cdot)$ is the fully connected layer, and $\sigma(\cdot)$ is the sigmoid function. Then, to predict video-level concepts, temporal attention aggregates all frame probabilities:
\begin{eqnarray}
& P^{v} = \sum_{t=1}^{T} \alpha_t p_t, \notag \\
& \alpha = \mathrm{softmax}(F W^{tp}), \label{eq.4-5}
\end{eqnarray}
where $P^{v} \in \mathbb{R}^{1 \times N_C}$ is the video concept probability, $W^{tp} \in \mathbb{R}^{d \times 1}$ is learnable parameter, $\alpha \in \mathbb{R}^{T}$ represents the temporal attention weights, and $\alpha_t$ is the weight at frame $t$.

Since only the video-level label $Y^C$ is available, frame-level concept detection is performed in a weakly supervised manner using multiple instance learning:
\begin{equation}
\mathcal{L}_{mil} = - \sum_{i=1}^{N_C} [Y^C_i \log P^{v}_i + (1-Y^C_i) \log (1 - P^{v}_i)], \label{eq.6}
\end{equation}
where $P^{v}_i$ represents the probability of the $i$-th concept.

Considering the high correlations between concepts and video events, the concept probabilities are weighted with learnable concept embeddings $W^{C} \in \mathbb{R}^{N_C \times d}$ to derive frame-level and video-level concepts:
\begin{equation}
f^{c}_t = p_t W^{C}, \, f^{vc} = P^{v} W^{C}, \label{eq.7-8}
\end{equation}
where $f^{vc} \in \mathbb{R}^{1 \times d}$ represents the video concept, and $f^{c}_t \in \mathbb{R}^{1 \times d}$ is the frame concept at time $t$. Then, each frame concept is combined with its corresponding frame feature:
\begin{equation}
\tilde{f}_t = f^{c}_t + f_t, \label{eq.9}
\end{equation}
where $\tilde{f}_t \in \mathbb{R}^{1 \times d}$ is the enhanced frame feature. The enhanced video features $\tilde{F} = \{\tilde{f}_t\}^T_{t=1}$ are used for event localization and caption generation.

\subsubsection{Concept Contrastive}
Given that concept detection is weakly supervised, contrastive learning is introduced to obtain discriminative concept embeddings, enabling the model to accurately distinguish various concepts in videos. For each video, positive and negative concepts are generated according to the ground-truth concept label $Y^C$. A positive concept is defined as one that has at least one intersection with the ground-truth label, while a negative concept has none. During training, $N_S$ pairs of positive and negative concept labels are sampled: $Y^{C^+}$ and $Y^{C^-}$. By combining the sampled labels and the concept embeddings $W^{C}$ as defined in Eq.~(\ref{eq.7-8}), we derive $N_S$ positive and negative concepts: $\{f^{vc^+}_n\}^{N_S}_{n=1}$ and $\{f^{vc^-}_n\}^{N_S}_{n=1}$. The triplet loss in the concept embedding space is defined as:
\begin{equation}
\mathcal{L}_{tri} = \frac{1}{N_S} \sum_{n=1}^{N_S} \left[\mathrm{sim}(f^{vc}, f^{vc^-}_n) - \mathrm{sim}(f^{vc}, f^{vc^+}_n) + \delta \right]_+, \label{eq.10}
\end{equation}
where $\left[\cdot\right]_+ = \mathrm{max}(0,\cdot)$, $\mathrm{sim}(\cdot,\cdot)$ is the cosine similarity, and $\delta$ denotes the margin. This loss encourages the model to learn concept embeddings such that the distance between the video concept and the positive concept is minimized, while the distance to the negative concept is maximized.

\subsection{Cyclic Co-learning}
\begin{figure}[t]
	\centering
	\includegraphics[width=0.45\textwidth]{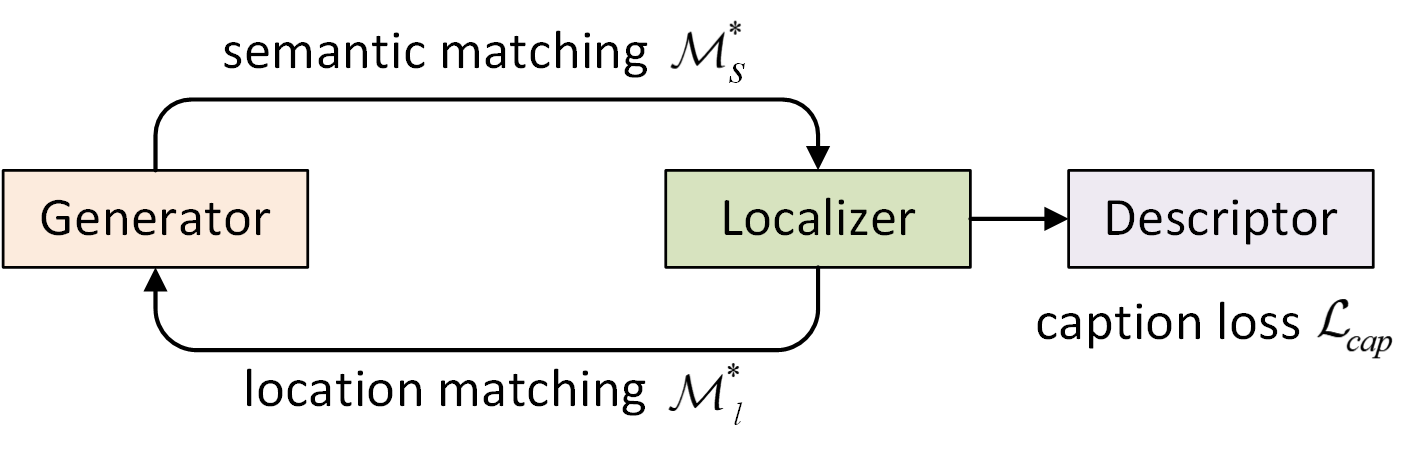}
	\caption{Cyclic co-learning.}
	\label{fig:colearning}
\end{figure}

The captioning network, based on PDVC~\cite{wang2021end}, consists of three components: a deformable transformer (generator), a localization head (localizer), and a captioning head (descriptor). The generator takes video features $\tilde{F}$ and $N$ learnable event queries $\{q_i\}^{N}_{i=1}$ as input, and outputs updated queries $\{\tilde{q}_i\}^{N}_{i=1}$ containing event semantics and temporal information. For each query $\tilde{q}_i$, the localizer predicts the event location $l_i$ (start time and end time), and the descriptor produces the caption $C_i = \{c_{i,1}, \dots, c_{i,L}\}$, where $L$ is the sentence length. As illustrated in Fig.\ref{fig:colearning}, the generator and localizer co-learn in a cycle: the generator perceives events and guides the localizer via semantic matching, while the localizer provides feedback through location matching to enhance the generator's semantic perception. Finally, the descriptor produces captions based on the queries selected by the localizer.

Given $N^*$ ground truth events $\{l^*_j, C^*_j\}^{N^*}_{j=1}$, where $l^*_j$ is the $j$-th event location and $C^*_j$ is the corresponding caption. To match the predicted events with ground truth in a global scheme, the localizer matches the predicted locations $\{l_i\}^{N}_{i=1}$ with the ground truth locations $\{l^*_j\}^{N^*}_{j=1}$ (location matching) using the Hungarian algorithm~\cite{kuhn1955hungarian}: 
\begin{equation}
\mathcal{M}^*_l = \mathrm{arg} \, \underset{\mathcal{M}}{\mathrm{min}} \sum_{(i,j) \in \mathcal{M}} \mathrm{cost}(l_i,l^*_j), \label{match:location}
\end{equation}
where $\mathcal{M}$ represents all possible matchings, and $\mathrm{cost}(\cdot,\cdot)$ is the gIOU~\cite{rezatofighi2019generalized} cost. The optimal location matching $\mathcal{M}^*_l= \{(i,j) | i=\pi(j), j \in \{1,\cdots, N^*\}\}$, where $\pi(j)$ returns the index of the predicted location that matches the ground truth location $l^*_j$. Based on the matching set $\mathcal{M}^*_l$, the localization-guided loss $\mathcal{L}_{lg}$ is defined:
\begin{equation}
\mathcal{L}_{lg} = \frac{1}{|\mathcal{M}^*_l|} \sum_{(i,j) \in \mathcal{M}^*_l} \left( \lambda_{1}\mathcal{L}_{giou}(l_i,l^*_j) + \lambda_{2}\mathcal{L}_{cap}(C_i,C^*_j) \right), \label{loss:lg}
\end{equation}
where $\lambda_{1}$ and $\lambda_{2}$ are hyperparameters, $\mathcal{L}_{giou}$ is the gIOU localization loss for localizer, and $\mathcal{L}_{cap}$ denotes the cross-entropy caption loss for descriptor. Obviously, the localizer selects queries for the descriptor based on location matching and improves the descriptor's captioning ability. However, without semantic guidance, the localizer may not effectively distinguish between semantically distinct nearby events. Thus, semantic matching is introduced in the generator:
\begin{equation}
\mathcal{M}^*_s = \mathrm{arg} \, \underset{\mathcal{M}}{\mathrm{max}} \sum_{(i,j) \in \mathcal{M}} \mathrm{sim}(\tilde{q}_i, z_j),\label{match:semantic}
\end{equation}
where $z_j$ is the semantic representation of $C^*_j$ from the CLIP text encoder, and $\mathrm{sim}(\cdot,\cdot)$ denotes the cosine similarity. To enhance the generator's semantic perception, the cosine distance between the queries selected by the localizer and the ground-truth semantics is minimized:
\begin{small}
	\begin{equation}
	\mathcal{L}_{sem} = \frac{1}{|\mathcal{M}^*_l|} \sum_{(i,j) \in \mathcal{M}^*_l} \left(1 - \mathrm{sim}(\tilde{q}_i, z_j) \right). \label{eq.sem}
	\end{equation}
\end{small}

Using the optimal semantic matching $\mathcal{M}^*_s$, the semantic-guided loss $\mathcal{L}_{sg}$ is defined as:
\begin{small}
	\begin{eqnarray}
	\notag \mathcal{L}_{sg} &=& \frac{1}{|\mathcal{M}^*_s|} \sum_{(i,j) \in \mathcal{M}^*_s} \left( \lambda_{1}\mathcal{L}_{giou}(l_i,l^*_j) + \lambda_{2}\mathcal{L}_{cap}(C_i,C^*_j) \right) \\
	&+& \lambda_{3}\mathcal{L}_{sem},\label{loss:sg}
	\end{eqnarray}
\end{small}
Semantic matching encourages the localizer and descriptor to perceive events from a semantic perspective. Nevertheless, it is sensitive to temporally distinct but semantically similar events (similar events recurring temporally). To address this, semantic and location matching are combined to derive the cyclic loss:
\begin{small}
	\begin{eqnarray}
	\mathcal{L}_{cyc} &=& \frac{1}{|\mathcal{M}^*_s|} \sum_{(i,j) \in \mathcal{M}^*_s} \lambda_{1}\mathcal{L}_{giou}(l_i,l^*_j) \notag \\ 
	&+& \frac{1}{|\mathcal{M}^*_l|} \sum_{(i,j) \in \mathcal{M}^*_l} \lambda_{2}\mathcal{L}_{cap}(C_i,C^*_j) \notag \\
	&+& \lambda_{3}\mathcal{L}_{sem}. \label{loss:cyc}
	\end{eqnarray}
\end{small}
In the cyclic loss $\mathcal{L}_{cyc}$, the generator improves the localizer's accuracy through semantic matching (the first term). The localizer, in turn, provides feedback to refine the generator's event semantic perception via location matching (the third term). This strategy enables mutual benefits between semantic perception and event localization.

\subsection{Training}
\label{sec:loss}
The overall loss is obtained:
\begin{equation}
\mathcal{L} = \mathcal{L}_{cyc} + \lambda_{4}\mathcal{L}_{tri} + \lambda_{5}\mathcal{L}_{mil},
\end{equation}
where $\lambda_{4}$, and $\lambda_{5}$ are hyperparameters.

\section{Experiment}

\begin{table}[t]
	\begin{center}
		\footnotesize
		\resizebox{\columnwidth}{!}{
			\setlength{\tabcolsep}{0.61mm}{
				\small
				\begin{tabular}{lcccccccc}
					\toprule[1.5pt]
					\multirow{2}{*}{Models} 
					& \multicolumn{4}{c}{Ground-truth proposals} & \multicolumn{4}{c}{Learned proposals} \\
					\cline{2-9}
					& B@4 & M & C & S & B@4 & M & C & S \\
					\midrule[0.5pt]
					\multicolumn{9}{l}{\emph{\footnotesize visual feature C3D}} \\
					HRNN 
					& 1.59 & 8.81 & 24.17 & - 
					& 0.70 & 5.68 & 12.35 & \\
					DCE
					& 1.60 & 8.88 & 25.12 & -
					& 0.71 & 5.69 & 12.43 & - \\
					DVC
					& 1.62 & 10.33 & 25.24 & - 
					& 0.73 & 6.93  & 12.61 & - \\
					GPaS
					& 1.53 & 11.04 & 28.20 & - 
					& 0.93 & 7.44  & 13.00 & - \\
					E2E-MT
					& 2.71 & 11.16 & 47.71 & - 
					& 1.15 & 4.98  & 9.25  & 4.02 \\
					PDVC$^\dag$
					& 2.74 & 10.34 & 48.55 & 8.94
					& 1.65 & 7.50  & 25.87 & 5.26 \\
					\midrule[0.5pt]
					\multicolumn{9}{l}{\emph{\footnotesize visual C3D + audio VGGish}} \\
					MDVC 
					& 1.98 & 11.07 & 45.39 & - 
					& 1.01 & 7.46  & 7.38  & - \\
					BMT 
					& 1.99 & 10.90 & 42.74 & - 
					& 1.88 & 8.44  & 11.35 & - \\
					GS-MS-FTN 
					& 1.91 & 10.93 & - & - 
					& 1.64 & 8.69  & - & - \\
					PDVC$^\dag$
					& 2.86 & 10.27 & 48.64 & 9.70
					& 1.96 & 8.08  & 28.59 & 5.42 \\
					\midrule[0.5pt]
					\multicolumn{9}{l}{\emph{\footnotesize visual feature CLIP}} \\
					Vid2Seq
					& - & - & - & -
					& - & 8.50 & 30.10 & 5.80 \\
					DIBS
					& - & - & - & -
					& - & 8.93 & 31.89 & 5.85 \\
					PDVC$^\dag$
					& 3.29 & 11.53 & 57.17 & 10.16
					& 1.75 & 7.63  & 30.22 & 6.16 \\
					CM$^{2\dag}$
					& 3.55 & 12.06 & 58.97 & 10.02
					& 2.29 & 8.51  & 32.89 & \textbf{6.21} \\
					\rowcolor{lightgray}
					MCCL 
					& \textbf{3.89} & \textbf{12.52} & \textbf{63.01} & \textbf{10.35}
					& \textbf{2.68} & \textbf{9.05}  & \textbf{34.92} & 6.16 \\
					\bottomrule[1.5pt]
				\end{tabular}
		}}
	\end{center}
	\caption{Caption results on ActivityNet Captions. B@4, M, C, S denote BLEU4, METEOR, CIDEr and SODA\_c, respectively. $\dag$ indicates reproduced from official code.}
	\label{tab:ActivityNet}
\end{table}

\subsection{Experimental Setup}

\subsubsection{Dataset} 
Experiments are conducted on two popular benchmark datasets, ActivityNet Captions~\cite{krishna2017dense} and YouCook2~\cite{zhou2018towards}, to demonstrate the effectiveness of the proposed method. ActivityNet Captions contains 20k untrimmed videos of human activities, averaging 120 seconds per video with 3.65 temporally localized sentences, totaling 100k sentences. We use the official split: 10,009/4,925/5,044 videos for training, validation, and testing. YouCook2 includes 2k untrimmed cooking videos, each averaging 320 seconds and annotated with 7.7 sentences. We follow the official split: 1,333/457/210 videos for training, validation, and testing.

\subsubsection{Evaluation Metrics} 
The method is evaluated from two aspects: 1) For dense captioning performance, evaluation tools from the ActivityNet Challenge 2018 are used, measuring CIDEr~\cite{vedantam2015cider}, BLEU4~\cite{papineni2002bleu}, and METEOR~\cite{banerjee2005meteor} scores. These metrics assess the average precision of the matched pairs between generated captions and ground truth at IOU thresholds of \{0.3, 0.5, 0.7, 0.9\}. Additionally, SODA\_c~\cite{fujita2020soda} is used to evaluate storytelling ability. 2) For localization performance, the harmonic mean of average precision and average recall at IOU thresholds of \{0.3, 0.5, 0.7, 0.9\}, along with the F1 score, is used.

\subsubsection{Implementation Details}
The experiments are developed with Python 3.9 and PyTorch 1.12 and performed on a single RTX 4090 GPU. For both datasets, each video is uniformly sampled or interpolated to $T$ frames, with $T$ set to 100 for ActivityNet Captions and 200 for YouCook2. CLIP ViT-L/14's image encoder extracts 768-dimensional frame features, while its text encoder extracts 768-dimensional sentence features from the corpus. For both datasets, the number of chunks is set to $W=20$, the number of retrieved sentences per chunk $N_K$ is set to 10. The number of concepts is set to $N_C=500$ for ActivityNet Captions and $N_C=600$ for YouCook2. The number of positive and negative samples $N_S$ is set to 10, with the margin $\delta$ is set to 0.5. The number of event queries is set to $N=10$ for ActivityNet Captions and $N=100$ for YouCook2. The hyperparameters $\lambda_{1}$, $\lambda_{2}$, $\lambda_{3}$, $\lambda_{4}$, and $\lambda_{5}$ are set to 4, 1, 0.5, 1, and 1, respectively.

\begin{table}[t]
	\begin{center}
		\resizebox{\columnwidth}{!}{
			\setlength{\tabcolsep}{0.71mm}{
				\large
				\begin{tabular}{lcccccccc}
					\toprule[1.5pt]
					\multirow{2}{*}{Models} & \multicolumn{4}{c}{Ground-truth proposals} & \multicolumn{4}{c}{Learned proposals} \\
					\cline{2-9}
					& B@4 & M & C & S  & B@4 & M & C & S\\
					\midrule[0.5pt]
					\multicolumn{9}{l}{\emph{\footnotesize Pretrain}} \\
					Vid2Seq & - & - & - & - & - & \textbf{9.30} & \textbf{47.10} & \textbf{7.90} \\ 
					DIBS    & - & - & - & - & - & 7.51 & 44.44 & 6.39 \\ 
					\midrule[0.5pt]
					\multicolumn{9}{l}{\emph{\footnotesize without Pretrain}} \\
					PDVC$^\dag$ & 3.79 & 14.05 & 86.39 & 16.38 & 1.62 & 5.72 & 30.82 & 5.03 \\ 
					CM$^{2\dag}$ & 3.56 & 14.09 & 89.43 & 15.10 & 1.64 & 5.56 & 30.15& 5.21 \\
					\rowcolor{lightgray}
					MCCL 		 & \textbf{4.69} & \textbf{14.69} & \textbf{96.95} & \textbf{16.85} & \textbf{2.04} & 6.53 & 36.09 & 5.21 \\
					\bottomrule[1.5pt]
				\end{tabular}
		}}
	\end{center}
	\caption{Caption results on YouCook2 dataset. $\dag$ denotes the results reproduced from official code.}
	\label{tab:YouCook2}
\end{table}

\begin{table}
	\begin{center}
		\resizebox{\columnwidth}{!}{
			\setlength{\tabcolsep}{1.85mm}{
				\large
				\begin{tabular}{l ccc  ccc}
					\toprule[1.5pt]
					& \multicolumn{3}{c}{Event Captions} & \multicolumn{3}{c}{Event Localization} \\
					\cline{2-7}
					& B@4 & M & C  & Recall & Precision & F1\\ 
					\midrule[0.5pt]
					PDVC$^\dag$			& 1.75 & 7.63  & 30.22 & 53.69 & 55.16 & 54.41\\
					CM$^{2\dag}$		& 2.29 & 8.51  & 32.89 & \textbf{53.73} & 56.14 & 54.91\\
					\hline
					MCCL($\mathcal{L}_{sg}$)  	& 1.36 & 6.75 & 23.43 & 46.50 & \textbf{61.04} & 52.79\\
					MCCL($\mathcal{L}_{lg}$)  	& 2.23 & 8.56 & 32.15 & 52.15 & 56.61 & 54.28\\
					MCCL($\mathcal{L}_{cyc}$) 	& \textbf{2.68} & \textbf{9.05}  & \textbf{34.92} & 53.19 & 57.36 & \textbf{55.23}\\
					\bottomrule[1.5pt]
				\end{tabular}
		}}
	\end{center}
	\caption{Localization performance on ActivityNet Captions. $\dag$ denotes the results reproduced from official code.}
	\label{tab:Localization}
\end{table}

\subsection{Experimental Results}
\begin{table*}[t]
	\small
	\begin{subtable}[t]{0.32\textwidth}
		\centering
		\setlength{\tabcolsep}{1.6 mm}{
			\begin{tabular}{c c c c c}
				\toprule[1.5pt]
				$N_K$  	  & B@4  & M  	 & C 	  & SODA\_c \\ 
				\midrule[0.5pt]
				5  & 2.41 & 8.84 & 32.03 & 5.97\\
				10 & \textbf{2.68} & \textbf{9.05} & 34.92 & 6.16\\
				15 & 2.61 & 8.93 & \textbf{35.21} & 6.14\\
				20 & 2.53 & 8.69 & 34.68 		  & 6.24\\
				30 & 2.56 & 8.87 & 34.90 		  & \textbf{6.34}\\
				50 & 2.52 & 8.71 & 34.32 		  & 6.28\\
				\bottomrule[1.5pt]
		\end{tabular}}
		\caption{Number of retrieved sentences $N_K$.}
		\label{tab:sentences}
	\end{subtable}
	{~}
	\begin{subtable}[t]{0.31\textwidth}
		\centering
		\setlength{\tabcolsep}{1.6 mm}{
			\begin{tabular}{c  c c c c}
				\toprule[1.5pt]
				$N_C$   & B@4         & M             & C             & SODA\_c \\ 
				\midrule[0.5pt]
				300 & 1.78 & 8.09 & 27.86 & 5.96\\
				400 & 2.41 & 8.65 & 33.02 & 6.12\\
				500 & 2.68 & \textbf{9.05} & \textbf{34.92} & \textbf{6.16}\\
				600 & \textbf{2.75} & 9.02 & 34.41 & 6.01\\
				700 & 2.62 & 8.81 & 32.53 & 6.06\\
				800 & 2.15 & 8.54 & 32.75 & 6.12\\
				\bottomrule[1.5pt]
		\end{tabular}}
		\caption{Number of concepts $N_C$.}
		\label{tab:concepts}
	\end{subtable}
	{~}
	\begin{subtable}[t]{0.35\textwidth}
		\centering
		\setlength{\tabcolsep}{1.6 mm}{
			\begin{tabular}{ c c c c c}
				\toprule[1.5pt]
				$N_S$   & B@4         & M             & C             & SODA\_c \\ 
				\midrule[0.5pt]
				0 & 2.01	      	& 8.25          & 30.11        		 & 5.96 \\ 
				5 & 2.42 			& 8.75 			& 34.69 			 & \textbf{6.20}\\
				10 & \textbf{2.68} 	& \textbf{9.05} & \textbf{34.92} 	& 6.16\\
				15 & 2.65 			& 8.89 		    & 34.83 			& 6.12\\
				20 & 2.66 			& 8.87			& 34.41 			& 6.14\\
				30 & 2.42 			& 8.60	& 33.73 			& 6.18\\
				\bottomrule[1.5pt]
		\end{tabular}}
		\caption{Number of positive and negative samples $N_S$.}
		\label{tab:samples}
	\end{subtable}
	\caption{Hyperparameter studies on the ActivityNet Captions dataset.}
	\label{tab:Hyperparameters}
\end{table*}

\begin{table}[t]
	\begin{center}
		\resizebox{0.97\columnwidth}{!}{
			\setlength{\tabcolsep}{1.61mm}{
				\begin{tabular}{ccc cccc}
					\toprule[1.5pt]
					V2T & MCD & CyC & B@4 & M & C & SODA\_c\\
					\midrule[0.5pt]
					\xmark & \xmark & \xmark  & 1.93	      & 7.68          & 27.74         & 5.61 \\ 
					\cmark & \xmark & \xmark  & 2.26	      & 8.54 		  & 31.32 		  & 6.01 \\
					\xmark & \cmark & \xmark  & 2.01	      & 8.03 		  & 29.33 		  & 5.91 \\
					\xmark & \xmark & \cmark  & 2.28	      & 8.39 		  & 30.98 		  & 5.96 \\
					\cmark & \cmark & \xmark  & 2.32 		  & 8.55 		  & 32.97 		  & \textbf{6.24}\\  
					\cmark & \cmark & \cmark  & \textbf{2.68} & \textbf{9.05} & \textbf{34.92} & 6.16\\
					\bottomrule[1.5pt]
				\end{tabular}
		}}
	\end{center}
	\caption{Performance of different components. V2T, MCD, and CyC denote video-to-text retrieval, multiple concept detection, and cyclic co-learning, respectively.}
	\label{tab:components}
\end{table}

We compare the performance of our method with that of state-of-the-art methods based on three types of approaches: 
(1) Methods using C3D features: HRNN~\cite{venugopalan2015sequence}, DCE~\cite{krishna2017dense}, DVC~\cite{li2018jointly}, GPaS~\cite{zhang2020dense}, E2E-MT~\cite{zhou2018end}, and PDVC~\cite{wang2021end}. 
(2) Methods using multimodal features: MDVC~\cite{iashin2020multi}, BMT~\cite{iashin2020better}, GS-MS-FTN~\cite{xie2023global}, and PDVC~\cite{wang2021end}. 
(3) Methods using CLIP features: PDVC~\cite{wang2021end}, Vid2Seq~\cite{yang2023vid2seq}, DIBS~\cite{wu2024dibs}, and CM$^2$~\cite{kim2024you}.

\subsubsection{Dense Captioning Performance}
As shown in Tab.\ref{tab:ActivityNet}, the performance of MCCL on ActivityNet Captions is reported. Models utilizing CLIP features as input outperform other methods, indicating that CLIP features provide superior semantic understanding. With ground-truth proposals, MCCL surpasses state-of-the-art models across all metrics, achieving a significant 4.04\% improvement in CIDEr. With learned proposals, despite a lower SODA\_c compared to CM$^2$, MCCL achieves the best results in BLEU4, METEOR, and CIDEr, showing a 2.03\% improvement in CIDEr.

Tab.\ref{tab:YouCook2} presents the performance of MCCL on YouCook2 dataset. Notably, Vid2seq~\cite{yang2023vid2seq} achieves the best performance by using an additional 15M videos for pretraining, and DIBS~\cite{wu2024dibs} shows sub-optimal results with an extra 56k videos for pretraining. In contrast, our method, similar to PDVC~\cite{wang2021end} and CM$^2$, does not involve any extra pretraining. With ground-truth proposals, MCCL outperforms competitors across all metrics, even achieving a notable 7.52\% improvement in CIDEr. With learned proposals, MCCL achieves the best results in BLEU4, METEOR, and CIDEr, demonstrating a 5.94\% improvement in CIDEr compared to CM$^2$.

\subsubsection{Event Localization Performance}
Tab.\ref{tab:Localization} compares MCCL with methods that use CLIP features as input on ActivityNet Captions datasets. We evaluate the impact of the cyclic loss ($\mathcal{L}_{cyc}$) by comparing it with the semantic-guided loss ($\mathcal{L}_{sg}$) and the localization-guided loss ($\mathcal{L}_{lg}$). $\mathcal{L}_{sg}$ achieves high precision but lower recall due to its reliance on semantic matching, making it challenging to handle semantically similar but temporally distinct events in complex videos. $\mathcal{L}_{lg}$, on the other hand, focuses on location information to identify temporal boundaries, significantly improving recall. This highlights the importance of location information in understanding complex videos. Finally, $\mathcal{L}_{cyc}$ enhances the interaction and optimization between the generator and localizer, helping the localizer identify temporal boundaries under semantic guidance and the generator perceive event semantics based on localizer's feedback. This mutual reinforcement of semantics and localization results in the best F1 score.

\subsection{Ablation Studies}
\begin{figure}[t]
	\centering
	\includegraphics[width=0.48\textwidth]{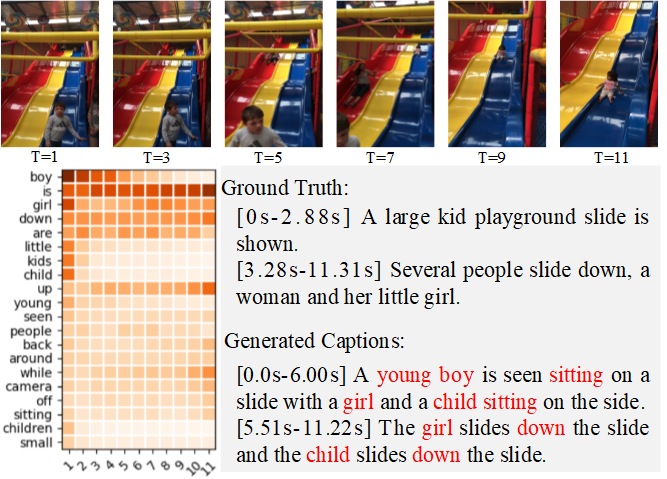}
	\caption{Concept guidance for video captioning.}
	\label{fig:illustrate_concepts}
\end{figure}

\begin{table}[t]
	\begin{center}
		\resizebox{0.95\columnwidth}{!}{
			\setlength{\tabcolsep}{2.2mm}{
				\small
				\begin{tabular}{l cccc}
					\toprule[1.5pt]
					& B@4         & M             & C             & SODA\_c \\ 
					\midrule[0.5pt]
					MCCL(w/o concept) & 2.57 & 8.76 & 32.46 & 5.71\\
					MCCL(w/ concept) & \textbf{2.68} & \textbf{9.05} & \textbf{34.92} & \textbf{6.16}\\
					\bottomrule[1.5pt]
				\end{tabular}
		}}
	\end{center}
	\caption{Effect of concept features on ActivityNet Captions.}
	\label{tab:concept}
\end{table}

\begin{figure*}[t]
	\centering
	\includegraphics[width=\textwidth]{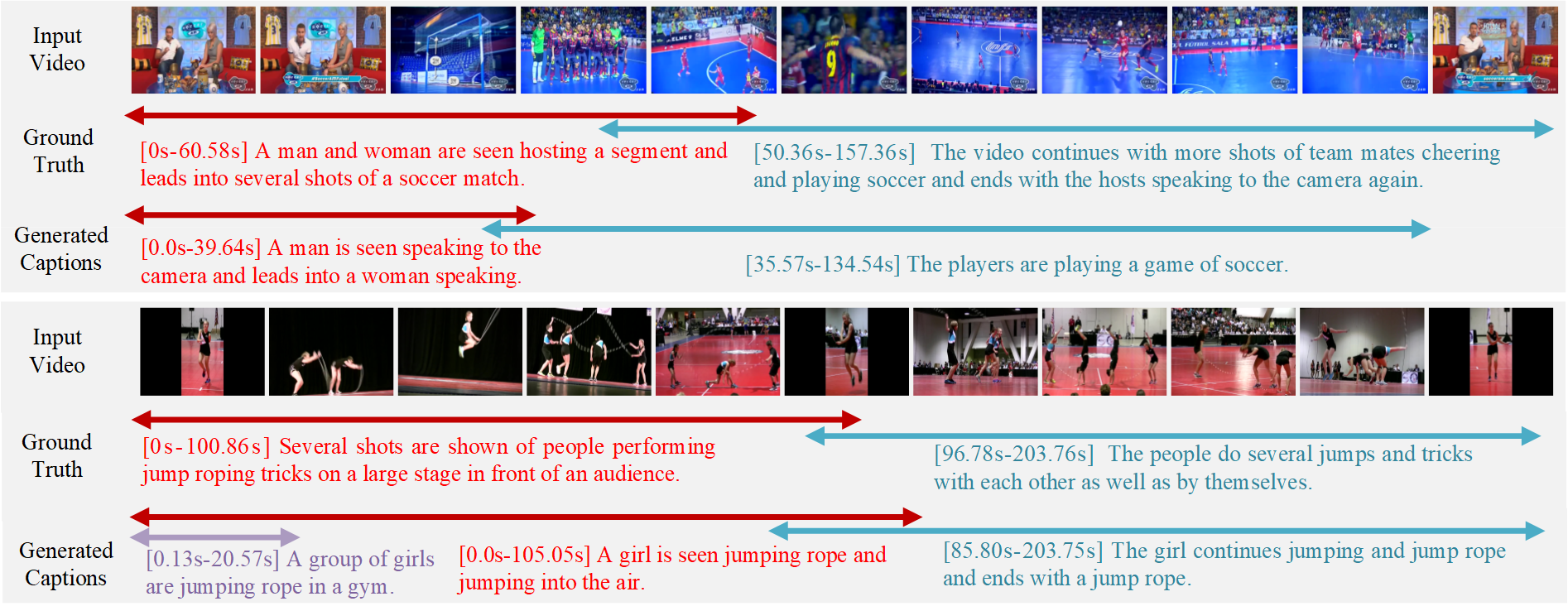}
	\caption{Qualitative captioning results. Two examples from ActivityNet Captions. The ground truth and generated captions for each video are shown separately.}
	\label{fig:caption_result}
\end{figure*}

\subsubsection{Ablation of Hyperparameters}
As shown in Tab.\ref{tab:sentences}, a small number of sentences is sufficient to achieve good performance. However, retrieving too many sentences results in higher SODA\_c, but lower METEOR and CIDEr scores. We hypothesize that a larger number of sentences may enrich semantics, which benefits SODA\_c, but also introduces more irrelevant information, which affects caption quality.

In Tab.\ref{tab:concepts}, the impact of the number of concepts on captioning performance is illustrated. It can be observed that an appropriate number of concepts improves the model's performance, whereas too many concepts may somewhat degrade caption quality.

Tab.\ref{tab:samples} shows the impact of concept contrastive learning. When the number of samples is set to $N_S=0$, it indicates that no contrastive learning is applied. The results show that concept contrastive learning significantly enhances performance by developing discriminative concepts, which help in identifying visual targets and provide cues for caption generation.

\subsubsection{Ablation Study of Different Components} 
Tab.\ref{tab:components} compares the functionality of different components. Through ablation studies on individual modules, it can be seen that integrating video-text retrieval (V2T) significantly enhances video captioning performance compared to the baseline without any components. This demonstrates the effectiveness of leveraging external knowledge to improve video understanding. While concept detection (MCD) shows limited standalone impact, its combination with V2T significantly improves all metrics, highlighting the crucial role of conceptual information in identifying visual targets and providing valuable event cues for caption generation. Cyclic co-learning (CyC) further enhances semantic perception and event localization, yielding notable gains in CIDEr and METEOR. Finally, integrating all modules achieves the best overall performance, despite a slight decrease in the SODA\_c metric.

\subsection{Qualitative Results}
\subsubsection{Effect of Concept Guidance} 
To see how concepts guide video captions, Fig.\ref{fig:illustrate_concepts} presents an example. The 11-second video uses the top 20 video-level concepts as reference, with frame-level probabilities visualized. MCCL identifies concepts such as `boy', `girl', and `child', as well as actions like `sitting' and `down', providing valuable temporal cues for understanding the video. Compared to the ground truth captions, our concept detection offers more fine-grained explanations. Although there is a misidentification of `woman' as `girl' and a lack of recognition of their relationship, we believe that discerning object relationships from visuals alone is challenging. Additional information, such as audio, could enhance video understanding. In addition, the comparison of experiments with and without incorporating concept features into video features is provided in Tab.\ref{tab:concept}. It is clear that MCCL with concept features shows improvements in all metrics.

\subsubsection{Captioning Results} 
Fig.\ref{fig:caption_result} shows two prediction examples from MCCL. It can be observed that our method effectively captures the temporal boundaries of events, leading to more accurate event localization. Additionally, the cyclic loss effectively handles complex videos, enhancing both event detection and caption quality.

\section{Conclusion}
In this paper, we propose a Multi-Concept Cyclic Learning (MCCL) network for dense video captioning, capable of performing frame-level multi-concept detection and enhancing video features to capture temporal event cues. Additionally, we introduce a cyclic co-learning strategy where the generator and localizer mutually reinforce  each other. Specifically, the generator uses semantic information to guide the localizer in event localization, while the localizer provides feedback through location matching to refine the generator's semantic perception. This mutual enhancement improves both semantic perception and event localization.

\section{Acknowledgements}
This work was supported by the National Natural Science Foundation of China (No.61976247).

\bibliography{aaai25}

\begin{thebibliography}{47}
\providecommand{\natexlab}[1]{#1}

\bibitem[{Aafaq et~al.(2022)Aafaq, Mian, Akhtar, Liu, and
  Shah}]{aafaq2022dense}
Aafaq, N.; Mian, A.; Akhtar, N.; Liu, W.; and Shah, M. 2022.
\newblock Dense video captioning with early linguistic information fusion.
\newblock \emph{IEEE Transactions on Multimedia}, 25: 2309--2322.

\bibitem[{Banerjee and Lavie(2005)}]{banerjee2005meteor}
Banerjee, S.; and Lavie, A. 2005.
\newblock METEOR: An automatic metric for MT evaluation with improved
  correlation with human judgments.
\newblock In \emph{Proceedings of the acl workshop on intrinsic and extrinsic
  evaluation measures for machine translation and/or summarization}, 65--72.

\bibitem[{Chang et~al.(2022)Chang, Zhao, Chen, Li, and Liu}]{chang2022event}
Chang, Z.; Zhao, D.; Chen, H.; Li, J.; and Liu, P. 2022.
\newblock Event-centric multi-modal fusion method for dense video captioning.
\newblock \emph{Neural Networks}, 146: 120--129.

\bibitem[{Chen et~al.(2023)Chen, Pan, Li, Yao, Chao, and
  Mei}]{chen2023retrieval}
Chen, J.; Pan, Y.; Li, Y.; Yao, T.; Chao, H.; and Mei, T. 2023.
\newblock Retrieval augmented convolutional encoder-decoder networks for video
  captioning.
\newblock \emph{ACM Transactions on Multimedia Computing, Communications and
  Applications}, 19(1s): 1--24.

\bibitem[{Chen and Jiang(2021)}]{chen2021towards}
Chen, S.; and Jiang, Y.-G. 2021.
\newblock Towards bridging event captioner and sentence localizer for weakly
  supervised dense event captioning.
\newblock In \emph{Proceedings of the IEEE/CVF Conference on Computer Vision
  and Pattern Recognition}, 8425--8435.

\bibitem[{Deng et~al.(2023)Deng, Chen, Qin, Chen, and Wu}]{deng2023prompt}
Deng, C.; Chen, Q.; Qin, P.; Chen, D.; and Wu, Q. 2023.
\newblock Prompt switch: Efficient clip adaptation for text-video retrieval.
\newblock In \emph{Proceedings of the IEEE/CVF International Conference on
  Computer Vision}, 15648--15658.

\bibitem[{Deng et~al.(2021)Deng, Chen, Chen, He, and Wu}]{deng2021sketch}
Deng, C.; Chen, S.; Chen, D.; He, Y.; and Wu, Q. 2021.
\newblock Sketch, ground, and refine: Top-down dense video captioning.
\newblock In \emph{Proceedings of the IEEE/CVF Conference on Computer Vision
  and Pattern Recognition}, 234--243.

\bibitem[{Duan et~al.(2018)Duan, Huang, Gan, Wang, Zhu, and
  Huang}]{duan2018weakly}
Duan, X.; Huang, W.; Gan, C.; Wang, J.; Zhu, W.; and Huang, J. 2018.
\newblock Weakly supervised dense event captioning in videos.
\newblock \emph{Advances in Neural Information Processing Systems}, 31.

\bibitem[{Fujita et~al.(2020)Fujita, Hirao, Kamigaito, Okumura, and
  Nagata}]{fujita2020soda}
Fujita, S.; Hirao, T.; Kamigaito, H.; Okumura, M.; and Nagata, M. 2020.
\newblock SODA: Story oriented dense video captioning evaluation framework.
\newblock In \emph{Computer Vision--ECCV 2020: 16th European Conference,
  Glasgow, UK, August 23--28, 2020, Proceedings, Part VI 16}, 517--531.
  Springer.

\bibitem[{Gao et~al.(2020)Gao, Wang, Song, and Liu}]{gao2020fused}
Gao, L.; Wang, X.; Song, J.; and Liu, Y. 2020.
\newblock Fused GRU with semantic-temporal attention for video captioning.
\newblock \emph{Neurocomputing}, 395: 222--228.

\bibitem[{Iashin and Rahtu(2020{\natexlab{a}})}]{iashin2020better}
Iashin, V.; and Rahtu, E. 2020{\natexlab{a}}.
\newblock A Better Use of Audio-Visual Cues: Dense Video Captioning with
  Bi-modal Transformer.
\newblock In \emph{The 31st British Machine Vision Virtual Conference}.

\bibitem[{Iashin and Rahtu(2020{\natexlab{b}})}]{iashin2020multi}
Iashin, V.; and Rahtu, E. 2020{\natexlab{b}}.
\newblock Multi-modal dense video captioning.
\newblock In \emph{Proceedings of the IEEE/CVF conference on computer vision
  and pattern recognition workshops}, 958--959.

\bibitem[{Jing et~al.(2023)Jing, Zhang, Zeng, Gao, Song, and
  Shen}]{jing2023memory}
Jing, S.; Zhang, H.; Zeng, P.; Gao, L.; Song, J.; and Shen, H.~T. 2023.
\newblock Memory-based augmentation network for video captioning.
\newblock \emph{IEEE Transactions on Multimedia}.

\bibitem[{Kim et~al.(2021)Kim, Kim, Lee, Park, and Kim}]{Kim2021viewpoint}
Kim, H.; Kim, J.; Lee, H.; Park, H.; and Kim, G. 2021.
\newblock Viewpoint-Agnostic Change Captioning with Cycle Consistency.
\newblock In \emph{Proceedings of the IEEE/CVF International Conference on
  Computer Vision}, 2075--2084.

\bibitem[{Kim et~al.(2024)Kim, Kim, Moon, Choi, and Kim}]{kim2024you}
Kim, M.; Kim, H.~B.; Moon, J.; Choi, J.; and Kim, S.~T. 2024.
\newblock Do You Remember? Dense Video Captioning with Cross-Modal Memory
  Retrieval.
\newblock In \emph{Proceedings of the IEEE/CVF Conference on Computer Vision
  and Pattern Recognition}, 13894--13904.

\bibitem[{Krishna et~al.(2017)Krishna, Hata, Ren, Fei-Fei, and
  Carlos~Niebles}]{krishna2017dense}
Krishna, R.; Hata, K.; Ren, F.; Fei-Fei, L.; and Carlos~Niebles, J. 2017.
\newblock Dense-captioning events in videos.
\newblock In \emph{Proceedings of the IEEE international conference on computer
  vision}, 706--715.

\bibitem[{Kuhn(1955)}]{kuhn1955hungarian}
Kuhn, H.~W. 1955.
\newblock The Hungarian method for the assignment problem.
\newblock \emph{Naval research logistics quarterly}, 2(1-2): 83--97.

\bibitem[{Li et~al.(2018)Li, Yao, Pan, Chao, and Mei}]{li2018jointly}
Li, Y.; Yao, T.; Pan, Y.; Chao, H.; and Mei, T. 2018.
\newblock Jointly localizing and describing events for dense video captioning.
\newblock In \emph{Proceedings of the IEEE conference on computer vision and
  pattern recognition}, 7492--7500.

\bibitem[{Lu et~al.(2024)Lu, Zhang, Yuan, Li, Wang, Li, and Hu}]{lu2024set}
Lu, Y.; Zhang, Z.; Yuan, C.; Li, P.; Wang, Y.; Li, B.; and Hu, W. 2024.
\newblock Set Prediction Guided by Semantic Concepts for Diverse Video
  Captioning.
\newblock In \emph{Proceedings of the AAAI Conference on Artificial
  Intelligence}, volume~38, 3909--3917.

\bibitem[{Luo et~al.(2022)Luo, Ji, Zhong, Chen, Lei, Duan, and
  Li}]{luo2022clip4clip}
Luo, H.; Ji, L.; Zhong, M.; Chen, Y.; Lei, W.; Duan, N.; and Li, T. 2022.
\newblock Clip4clip: An empirical study of clip for end to end video clip
  retrieval and captioning.
\newblock \emph{Neurocomputing}, 508: 293--304.

\bibitem[{Mun et~al.(2019)Mun, Yang, Ren, Xu, and Han}]{mun2019streamlined}
Mun, J.; Yang, L.; Ren, Z.; Xu, N.; and Han, B. 2019.
\newblock Streamlined dense video captioning.
\newblock In \emph{Proceedings of the IEEE/CVF conference on computer vision
  and pattern recognition}, 6588--6597.

\bibitem[{Papineni et~al.(2002)Papineni, Roukos, Ward, and
  Zhu}]{papineni2002bleu}
Papineni, K.; Roukos, S.; Ward, T.; and Zhu, W.-J. 2002.
\newblock Bleu: a method for automatic evaluation of machine translation.
\newblock In \emph{Proceedings of the 40th annual meeting of the Association
  for Computational Linguistics}, 311--318.

\bibitem[{Radford et~al.(2021)Radford, Kim, Hallacy, Ramesh, Goh, Agarwal,
  Sastry, Askell, Mishkin, Clark et~al.}]{radford2021learning}
Radford, A.; Kim, J.~W.; Hallacy, C.; Ramesh, A.; Goh, G.; Agarwal, S.; Sastry,
  G.; Askell, A.; Mishkin, P.; Clark, J.; et~al. 2021.
\newblock Learning transferable visual models from natural language
  supervision.
\newblock In \emph{International conference on machine learning}, 8748--8763.
  PMLR.

\bibitem[{Rahman, Xu, and Sigal(2019)}]{rahman2019watch}
Rahman, T.; Xu, B.; and Sigal, L. 2019.
\newblock Watch, listen and tell: Multi-modal weakly supervised dense event
  captioning.
\newblock In \emph{Proceedings of the IEEE/CVF international conference on
  computer vision}, 8908--8917.

\bibitem[{Rezatofighi et~al.(2019)Rezatofighi, Tsoi, Gwak, Sadeghian, Reid, and
  Savarese}]{rezatofighi2019generalized}
Rezatofighi, H.; Tsoi, N.; Gwak, J.; Sadeghian, A.; Reid, I.; and Savarese, S.
  2019.
\newblock Generalized intersection over union: A metric and a loss for bounding
  box regression.
\newblock In \emph{Proceedings of the IEEE/CVF conference on computer vision
  and pattern recognition}, 658--666.

\bibitem[{Ryu et~al.(2021)Ryu, Kang, Kang, and Yoo}]{ryu2021semantic}
Ryu, H.; Kang, S.; Kang, H.; and Yoo, C.~D. 2021.
\newblock Semantic grouping network for video captioning.
\newblock In \emph{proceedings of the AAAI Conference on Artificial
  Intelligence}, volume~35, 2514--2522.

\bibitem[{Shen et~al.(2017)Shen, Li, Su, Li, Chen, Jiang, and
  Xue}]{shen2017weakly}
Shen, Z.; Li, J.; Su, Z.; Li, M.; Chen, Y.; Jiang, Y.-G.; and Xue, X. 2017.
\newblock Weakly supervised dense video captioning.
\newblock In \emph{Proceedings of the IEEE Conference on Computer Vision and
  Pattern Recognition}, 1916--1924.

\bibitem[{Suin and Rajagopalan(2020)}]{suin2020efficient}
Suin, M.; and Rajagopalan, A. 2020.
\newblock An efficient framework for dense video captioning.
\newblock In \emph{Proceedings of the AAAI Conference on Artificial
  Intelligence}, volume~34, 12039--12046.

\bibitem[{Tian, Hu, and Xu(2021)}]{tian2021cyclic}
Tian, Y.; Hu, D.; and Xu, C. 2021.
\newblock Cyclic co-learning of sounding object visual grounding and sound
  separation.
\newblock In \emph{Proceedings of the IEEE/CVF Conference on Computer Vision
  and Pattern Recognition}, 2745--2754.

\bibitem[{Tu et~al.(2023)Tu, Li, Su, Zha, Yan, and Huang}]{tu2023self}
Tu, Y.; Li, L.; Su, L.; Zha, Z.-J.; Yan, C.; and Huang, Q. 2023.
\newblock Self-supervised cross-view representation reconstruction for change
  captioning.
\newblock In \emph{Proceedings of the IEEE/CVF International Conference on
  Computer Vision}, 2805--2815.

\bibitem[{Tu et~al.(2021)Tu, Li, Yan, Gao, and Yu}]{tu2021r}
Tu, Y.; Li, L.; Yan, C.; Gao, S.; and Yu, Z. 2021.
\newblock Rˆ3Net: Relation-embedded Representation Reconstruction Network for
  Change Captioning.
\newblock In \emph{Proceedings of the 2021 Conference on Empirical Methods in
  Natural Language Processing}, 9319--9329.

\bibitem[{Vedantam, Lawrence~Zitnick, and Parikh(2015)}]{vedantam2015cider}
Vedantam, R.; Lawrence~Zitnick, C.; and Parikh, D. 2015.
\newblock Cider: Consensus-based image description evaluation.
\newblock In \emph{Proceedings of the IEEE conference on computer vision and
  pattern recognition}, 4566--4575.

\bibitem[{Venugopalan et~al.(2015)Venugopalan, Rohrbach, Donahue, Mooney,
  Darrell, and Saenko}]{venugopalan2015sequence}
Venugopalan, S.; Rohrbach, M.; Donahue, J.; Mooney, R.; Darrell, T.; and
  Saenko, K. 2015.
\newblock Sequence to sequence-video to text.
\newblock In \emph{Proceedings of the IEEE international conference on computer
  vision}, 4534--4542.

\bibitem[{Wang et~al.(2018)Wang, Jiang, Ma, Liu, and
  Xu}]{wang2018bidirectional}
Wang, J.; Jiang, W.; Ma, L.; Liu, W.; and Xu, Y. 2018.
\newblock Bidirectional attentive fusion with context gating for dense video
  captioning.
\newblock In \emph{Proceedings of the IEEE conference on computer vision and
  pattern recognition}, 7190--7198.

\bibitem[{Wang et~al.(2021)Wang, Zhang, Lu, Zheng, Cheng, and
  Luo}]{wang2021end}
Wang, T.; Zhang, R.; Lu, Z.; Zheng, F.; Cheng, R.; and Luo, P. 2021.
\newblock End-to-end dense video captioning with parallel decoding.
\newblock In \emph{Proceedings of the IEEE/CVF International Conference on
  Computer Vision}, 6847--6857.

\bibitem[{Wang et~al.(2020)Wang, Zheng, Yu, Tian, and Hu}]{wang2020event}
Wang, T.; Zheng, H.; Yu, M.; Tian, Q.; and Hu, H. 2020.
\newblock Event-centric hierarchical representation for dense video captioning.
\newblock \emph{IEEE Transactions on Circuits and Systems for Video
  Technology}, 31(5): 1890--1900.

\bibitem[{Wu et~al.(2023)Wu, Liu, Huang, Bao, Xi, and Yu}]{wu2023concept}
Wu, B.; Liu, B.; Huang, P.; Bao, J.; Xi, P.; and Yu, J. 2023.
\newblock Concept parser with multimodal graph learning for video captioning.
\newblock \emph{IEEE Transactions on Circuits and Systems for Video
  Technology}, 33(9): 4484--4495.

\bibitem[{Wu et~al.(2024)Wu, Liu, Qiao, and Sun}]{wu2024dibs}
Wu, H.; Liu, H.; Qiao, Y.; and Sun, X. 2024.
\newblock DIBS: Enhancing Dense Video Captioning with Unlabeled Videos via
  Pseudo Boundary Enrichment and Online Refinement.
\newblock In \emph{Proceedings of the IEEE/CVF Conference on Computer Vision
  and Pattern Recognition}, 18699--18708.

\bibitem[{Xie et~al.(2023{\natexlab{a}})Xie, Sun, Xiong, Zheng, Zhao, and
  Zhou}]{xie2023ra}
Xie, C.-W.; Sun, S.; Xiong, X.; Zheng, Y.; Zhao, D.; and Zhou, J.
  2023{\natexlab{a}}.
\newblock Ra-clip: Retrieval augmented contrastive language-image pre-training.
\newblock In \emph{Proceedings of the IEEE/CVF Conference on Computer Vision
  and Pattern Recognition}, 19265--19274.

\bibitem[{Xie et~al.(2023{\natexlab{b}})Xie, Niu, Zhang, and
  Ren}]{xie2023global}
Xie, Y.; Niu, J.; Zhang, Y.; and Ren, F. 2023{\natexlab{b}}.
\newblock Global-shared Text Representation based Multi-Stage Fusion
  Transformer Network for Multi-modal Dense Video Captioning.
\newblock \emph{IEEE Transactions on Multimedia}.

\bibitem[{Yang et~al.(2023)Yang, Nagrani, Seo, Miech, Pont-Tuset, Laptev,
  Sivic, and Schmid}]{yang2023vid2seq}
Yang, A.; Nagrani, A.; Seo, P.~H.; Miech, A.; Pont-Tuset, J.; Laptev, I.;
  Sivic, J.; and Schmid, C. 2023.
\newblock Vid2seq: Large-scale pretraining of a visual language model for dense
  video captioning.
\newblock In \emph{Proceedings of the IEEE/CVF Conference on Computer Vision
  and Pattern Recognition}, 10714--10726.

\bibitem[{Yang, Cao, and Zou(2023)}]{yang2023concept}
Yang, B.; Cao, M.; and Zou, Y. 2023.
\newblock Concept-aware video captioning: Describing videos with effective
  prior information.
\newblock \emph{IEEE Transactions on Image Processing}.

\bibitem[{Yue et~al.(2024)Yue, Tu, Li, Gao, and Yu}]{yue2024multi}
Yue, S.; Tu, Y.; Li, L.; Gao, S.; and Yu, Z. 2024.
\newblock Multi-grained Representation Aggregating Transformer with Gating
  Cycle for Change Captioning.
\newblock \emph{ACM Transactions on Multimedia Computing, Communications and
  Applications}.

\bibitem[{Zhang et~al.(2021)Zhang, Qi, Yuan, Shan, Li, Deng, and
  Hu}]{zhang2021open}
Zhang, Z.; Qi, Z.; Yuan, C.; Shan, Y.; Li, B.; Deng, Y.; and Hu, W. 2021.
\newblock Open-book video captioning with retrieve-copy-generate network.
\newblock In \emph{Proceedings of the IEEE/CVF conference on computer vision
  and pattern recognition}, 9837--9846.

\bibitem[{Zhang et~al.(2020)Zhang, Xu, Ouyang, and Zhou}]{zhang2020dense}
Zhang, Z.; Xu, D.; Ouyang, W.; and Zhou, L. 2020.
\newblock Dense video captioning using graph-based sentence summarization.
\newblock \emph{IEEE Transactions on Multimedia}, 23: 1799--1810.

\bibitem[{Zhou, Xu, and Corso(2018)}]{zhou2018towards}
Zhou, L.; Xu, C.; and Corso, J. 2018.
\newblock Towards automatic learning of procedures from web instructional
  videos.
\newblock In \emph{Proceedings of the AAAI Conference on Artificial
  Intelligence}, volume~32.

\bibitem[{Zhou et~al.(2018)Zhou, Zhou, Corso, Socher, and Xiong}]{zhou2018end}
Zhou, L.; Zhou, Y.; Corso, J.~J.; Socher, R.; and Xiong, C. 2018.
\newblock End-to-end dense video captioning with masked transformer.
\newblock In \emph{Proceedings of the IEEE conference on computer vision and
  pattern recognition}, 8739--8748.

\end{thebibliography}

\newpage

\appendix
\section{Appendix A: Parallel Decoding}
This paper builds upon PDVC~\cite{wang2021end}, which consists of a deformable Transformer (generator), a localization head (localizer), a captioning head (descriptor), and an event counter operating in parallel. The deformable Transformer takes video features $\tilde{F}$ and $N$ learnable event queries $\{q_i\}^{N}_{i=1}$ as input and outputs queries $\{\tilde{q}_i\}^{N}_{i=1}$ containing semantic and temporal information of events.

\textbf{Localization Head}. The localizer predicts a temporal location and foreground confidence for each event query $\tilde{q}_i$. Both implemented through a multi-layer perceptron. Localizer outputs a set of tuples ${(t^s_i, t^e_i, c_i)}^{N}_{i=1}$, where each tuple represents the start time $t^s_i$, end time $t^e_i$, and confidence score $a_i$ for each query $\tilde{q}_i$. The localizer is optimized by localization loss $\mathcal{L}_{giou}$, which measures the gIOU between the predicted and ground-truth locations.

\textbf{Captioning Head}. The generator $\mathcal{G}$ is implemented using a deformable soft attention (DSA)~\cite{wang2021end}. When generating the $l$-th word $w_{i,l}$ for $i$-th query $\tilde{q}_i$, the DSA uses soft attention around the reference point $\tilde{p}_i$ to extract a local feature $z_{i,l}$ from the video features $\tilde{F}$. The local feature $z_{i,l}$, along with event query $\tilde{q}_i$ and previous word $w_{i,l-1}$, are then fed into an LSTM, generating the next word: $w_{i,l} = \mathcal{G}(\tilde{q}_i, z_{i,l}, w_{i,l-1})$. The reference point $\tilde{p}_i$ is determined by both the LSTM's hidden state $h_{i,l}$ and the query $\tilde{q}_i$. This process continues iteratively, generating sentence for query $\tilde{q}_i$: $W_i = \{w_{i,1}, \dots, w_{i,L}\}$, where $L$ is the sentence length. The caption loss $\mathcal{L}_{cap}$ is defined as the cross-entropy between the predicted word probabilities and the ground truth.

\textbf{Event Counter}. The counter predicts a vector $r_{len}$ for each event query $\tilde{q}_i$, where each value in $r_{len}$ represents the likelihood of a specific number. During inference, the predicted number of events is obtained: $N_{set} = argmax(r_{len})$. The count loss $\mathcal{L}_{ct}$ is cross-entropy between the predicted count distribution and the ground truth.

\textbf{Set Loss}. For the predicted event locations and captions, the Hungarian algorithm is employed to find the optimal bipartite matching between predicted events and ground-truth events. The matching cost is: $\mathcal{L}_{giou}$ + $\mathcal{L}_{cls}$, where $\mathcal{L}_{cls}$ denotes the focal loss between the predicted classification scores and the ground-truth labels. The matched pairs are then used to compute the set loss $\mathcal{L}_{set}$:
\begin{equation}
\mathcal{L}_{set} = \beta_{giou}\mathcal{L}_{giou} + \beta_{cls}\mathcal{L}_{cls} + \beta_{cap}\mathcal{L}_{cap} + \beta_{ct}\mathcal{L}_{ct}. \label{eq.set}
\end{equation}
where $\beta_{giou}$, $\beta_{cls}$, $\beta_{cap}$ and $\beta_{ct}$ are hyperparameters.

\begin{table}[h]
	\begin{center}
		\resizebox{0.85\columnwidth}{!}{
			\setlength{\tabcolsep}{3.2mm}{
				\small
				\begin{tabular}{l cccc}
					\toprule[1.5pt]
					$N_K$   & B@4         & M             & C             & SODA\_c \\  
					\midrule[0.5pt]
					5  & 1.81 & 5.84 & 31.46 & 4.45\\
					10 & \textbf{2.04} & \textbf{6.53} & \textbf{36.09} & \textbf{5.21}\\
					15 & 1.97 & 6.28 & 35.39 & 5.16\\
					20 & 2.03 & 5.96 & 33.90 & 5.19\\
					30 & 1.94 & 6.12 & 34.09 & 5.13\\
					\bottomrule[1.5pt]
				\end{tabular}
		}}
	\end{center}
	\caption{Number of retrieved sentences $N_K$.}
	\label{tab:7}
\end{table}
\begin{table}[h]
	\begin{center}
		\resizebox{0.85\columnwidth}{!}{
			\setlength{\tabcolsep}{3.2mm}{
				\small
				\begin{tabular}{l cccc}
					\toprule[1.5pt]
					$N_C$   & B@4         & M             & C             & SODA\_c \\ 
					\midrule[0.5pt]
					400 & 1.92 & 6.13 & 32.94 & 4.86\\
					500 & 1.96 & 6.34 & 35.64 & \textbf{5.28}\\
					600 & 2.04 & \textbf{6.53} & \textbf{36.09} & 5.21\\
					700 & \textbf{2.17} & 6.35 & 35.67 & 5.15\\
					800 & 1.97 & 6.09 & 32.93 & 5.06\\
					\bottomrule[1.5pt]
				\end{tabular}
		}}
	\end{center}
	\caption{Number of concepts $N_C$.}
	\label{tab:8}
\end{table}
\begin{table}[h]
	\begin{center}
		\resizebox{0.85\columnwidth}{!}{
			\setlength{\tabcolsep}{3.2mm}{
				\small
				\begin{tabular}{l cccc}
					\toprule[1.5pt]
					$N_S$   & B@4         & M             & C             & SODA\_c \\ 
					\midrule[0.5pt]
					0 & 1.67 & 5.65 & 31.49 & 4.10\\
					5 & 2.00 & 6.13 & 33.71 & 4.76\\
					10 &  2.04 & \textbf{6.53} & \textbf{36.09} & \textbf{5.21}\\
					15 & \textbf{2.11} & 6.01 & 35.40 & 5.08\\
					20 & 1.92 & 6.20 & 34.83 & 5.14\\
					\bottomrule[1.5pt]
				\end{tabular}
		}}
	\end{center}
	\caption{Number of positive and negative samples $N_S$.}
	\label{tab:9}
\end{table}
\section{Appendix B: Ablation Results}
\subsubsection{Hyperparameters Validation on YouCook2} 
In the main paper, we demonstrate that selecting appropriate hyperparameters significantly enhances model performance on ActivityNet Captions. Tab.\ref{tab:7}, Tab.\ref{tab:8}, and Tab.\ref{tab:9} respectively verify the impact of hyperparameters $N_K$, $N_C$ and $N_S$ on model performance for the YouCook2 dataset.

\subsubsection{Visualization of Concept Detection} 
\begin{figure}[h]
	\centering
	\includegraphics[width=0.4\textwidth]{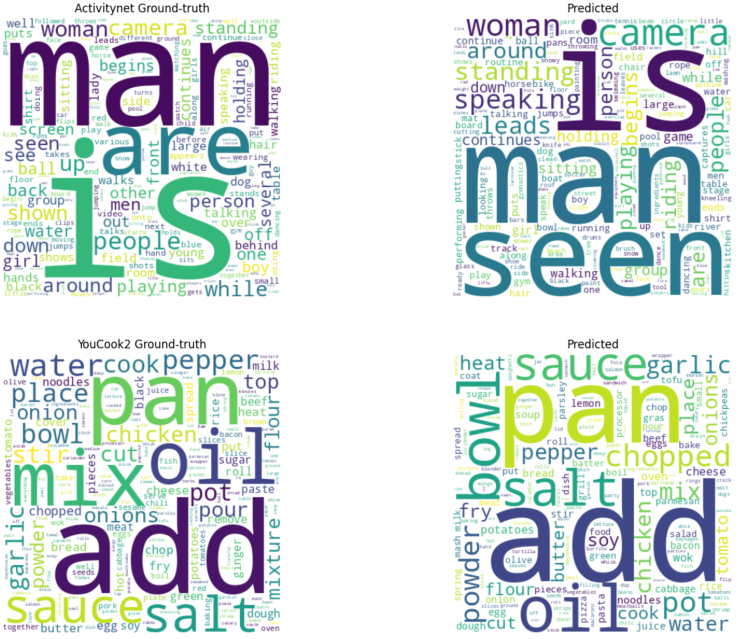}
	\caption{Representative concept words. The larger the font, the higher the frequency.}
	\label{fig:wordcloud}
\end{figure}

In Fig.\ref{fig:wordcloud}, we visualize representative concept words from the ActivityNet Captions and YouCook2 test sets. The left column shows the ground truth concept word statistics, while the right column displays the video-level concept word predictions from our model. The results indicate that the multiple concept detection module effectively captures concept words within videos, providing valuable cues for event localization and captioning.

\begin{table*}[t]
	\begin{center}
		\resizebox{0.95\textwidth}{!}{
			\setlength{\tabcolsep}{1.61mm}{
				\begin{tabular}{ c | c c c c c | c c c c c | c}
					\toprule[1.5pt]
					& \multicolumn{5}{c}{Recall} & \multicolumn{5}{c}{Precision} & F1\\ 
					Methods & 0.3 & 0.5 & 0.7 & 0.9 & avg & 0.3 & 0.5 & 0.7 & 0.9 & avg &  \\
					\midrule[0.5pt]
					PDVC$^\dag$ & \textbf{88.74} & \textbf{69.58} & 42.35 & 14.08 & 53.69 & 94.90 & 72.69 & 40.30 & 12.74 & 55.16 & 54.41\\
					CM$^2$$^\dag$ & 87.04 & 67.92 & 43.08 & \textbf{16.86} & \textbf{53.73} & 97.28 & 75.04 & 38.12 & 14.13 & 56.14 & 54.91\\
					\midrule[0.5pt]
					MCCL($\mathcal{L}_{sg}$)  & 79.86 & 56.71 & 35.74 & 13.70 & 46.50  & \textbf{99.53} & \textbf{85.50} & \textbf{45.67} & 13.15 & \textbf{61.04} & 52.79 \\
					MCCL($\mathcal{L}_{lg}$)  & 86.46 & 66.41 & 40.62 & 15.12 & 52.15 & 95.28 & 75.37 & 42.13 & 13.65 & 56.61 & 54.28 \\
					MCCL($\mathcal{L}_{cyc}$)  & 87.12 & 66.36 & \textbf{43.21} & 16.08 & 53.19 & 96.34 & 77.56 & 41.06 & \textbf{14.46} & 57.36 & \textbf{55.23} \\\bottomrule[1.5pt]
				\end{tabular}
		}}
	\end{center}
	\caption{Event localization on ActivityNet Captions validation set.}
	\label{tab:10}
\end{table*}

\subsubsection{Performance of Event Localization} In the main paper, we compare the performance of the semantic-guided loss $\mathcal{L}_{sg}$, localization-guided loss $\mathcal{L}_{lg}$, and cyclic loss $\mathcal{L}_{cyc}$. Tab.\ref{tab:10} provides detailed localization performance metrics, including average recall, average precision at IOU thresholds of \{0.3, 0.5, 0.7, 0.9\}, and their harmonic mean, F1 score.

$\mathcal{L}_{sg}$ demonstrates exceptional precision, particularly at lower IOU thresholds (0.3 and 0.5). Its average precision score of 61.04 highlights its effectiveness in maintaining semantic consistency. However, this precision often comes at the cost of lower recall, indicating potential missed events.

$\mathcal{L}_{lg}$ achieves a good balance between recall and precision. The F1 score of 54.28 reflects solid performance in both localization and description. Although it may not match the precision of $\mathcal{L}_{sg}$, it provides a balanced trade-off between precision and recall.

$\mathcal{L}_{cyc}$ exhibits the most balanced performance overall, with the highest F1 score of 55.23. By integrating both semantic and location information through a cyclic mechanism, this loss function optimally balances precision and recall.

\subsubsection{Computational complexity} In Tab.\ref{tab:11}, we compare PDVC, CM$^2$, and MCCL in terms of parameters, training/testing times, GFLOPs, and GPU memory usage. All experiments are conducted on the same device with the batch size set to 1. The results show that MCCL outperforms CM$^2$ in time, FLOPs, and memory usage.

\begin{table}[h]
	\begin{center}
		\resizebox{0.98\columnwidth}{!}{
			\setlength{\tabcolsep}{1.2mm}{
				\small
				\begin{tabular}{l cccc}
					\toprule[1.5pt]
					Method  & Parameters   &  Training/Testing  & FLOPs  & GPU memory \\  
					& (M)   & time(s)  &  (G) & usage(G) \\ 
					\midrule[0.5pt]
					PDVC & 18.03 & 0.050/0.027 & 12.64 & 2.63 \\
					CM$^2$ & 78.89 & 0.073/0.048 & 37.45 & 4.32\\
					MCCL & 96.31 & 0.065/0.042 & 31.90 & 4.03\\
					\bottomrule[1.5pt]
				\end{tabular}
		}}
	\end{center}
	\caption{Complexity comparison.}
	\label{tab:11}
\end{table}

\subsubsection{Number of Chunks} Tab.\ref{tab:12} presents the performance and computational complexity under different chunk settings. It can be observed that an excessive number of chunks noticeably increases the time overhead, while offering only slight improvements in performance.

\begin{table}[h]
	\begin{center}
		\resizebox{0.98\columnwidth}{!}{
			\setlength{\tabcolsep}{3.2mm}{
				\small
				\begin{tabular}{l ccccc}
					\toprule[1.5pt]
					$W$   & B@4         & M             & C    & Training/Testing  & FLOPs\\ 
					 &        &           &    & time(s)  & (G)\\
					\midrule[0.5pt]
					10 & 2.18 & 8.57 & 33.37 & 0.054/0.035 & 31.84 \\
					20 & 2.68 & 9.05 & 34.92 & 0.065/0.042 & 31.90\\
					30 & 2.55 & 8.73 & 34.58 & 0.079/0.055 & 31.97 \\
					40 & 2.53 & 8.82 & 35.14 & 0.092/0.069 & 32.04 \\
					50 & 2.36 & 8.67 & 34.62 & 0.107/0.083 & 32.12 \\
					\bottomrule[1.5pt]
				\end{tabular}
		}}
	\end{center}
	\caption{Different chunk settings.}
	\label{tab:12}
\end{table}

\section{Appendix C: Additional Qualitative Results}
\begin{figure*}[h]
	\centering
	\includegraphics[width=\textwidth]{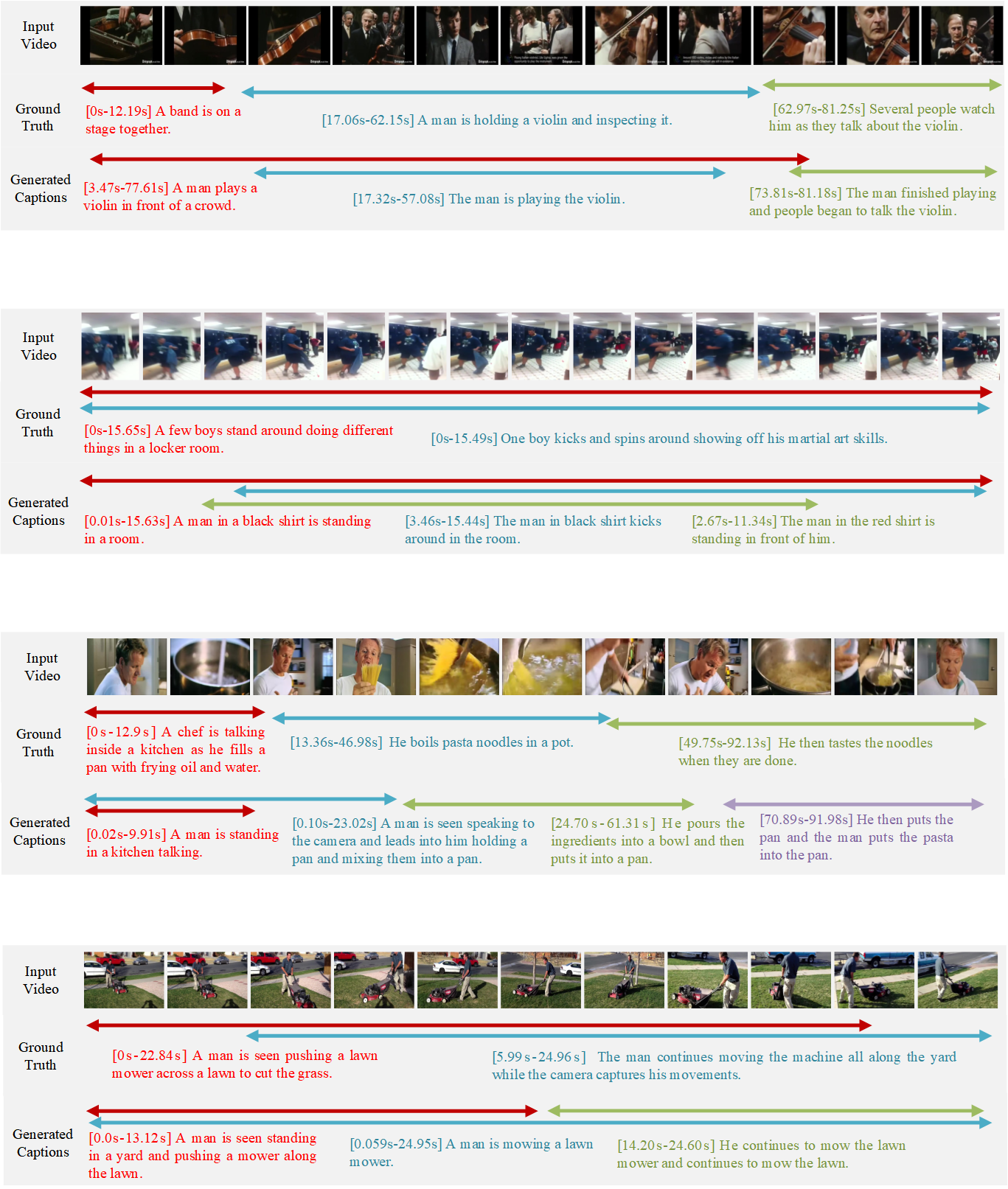}
	\caption{Examples of video captioning and localization.}
	\label{fig:caption_results_supp}
\end{figure*}

In the main paper, we demonstrate that concept detection can effectively guide video captioning. Fig.\ref{fig:illustrate_concepts_supp} provides additional examples, where the ground truth captions, predicted captions, and frame-level concept probabilities are listed for each video. The left side of each sample shows the top 20 predicted video-level concepts, while the horizontal axis represents time. Furthermore, Fig.\ref{fig:caption_results_supp} presents examples of video captioning, highlighting the performance in both caption generation and localization. These results show the interpretability and effectiveness of our approach.

\subsubsection{Limitations} 
\begin{figure*}[h]
	\centering
	\includegraphics[width=\textwidth]{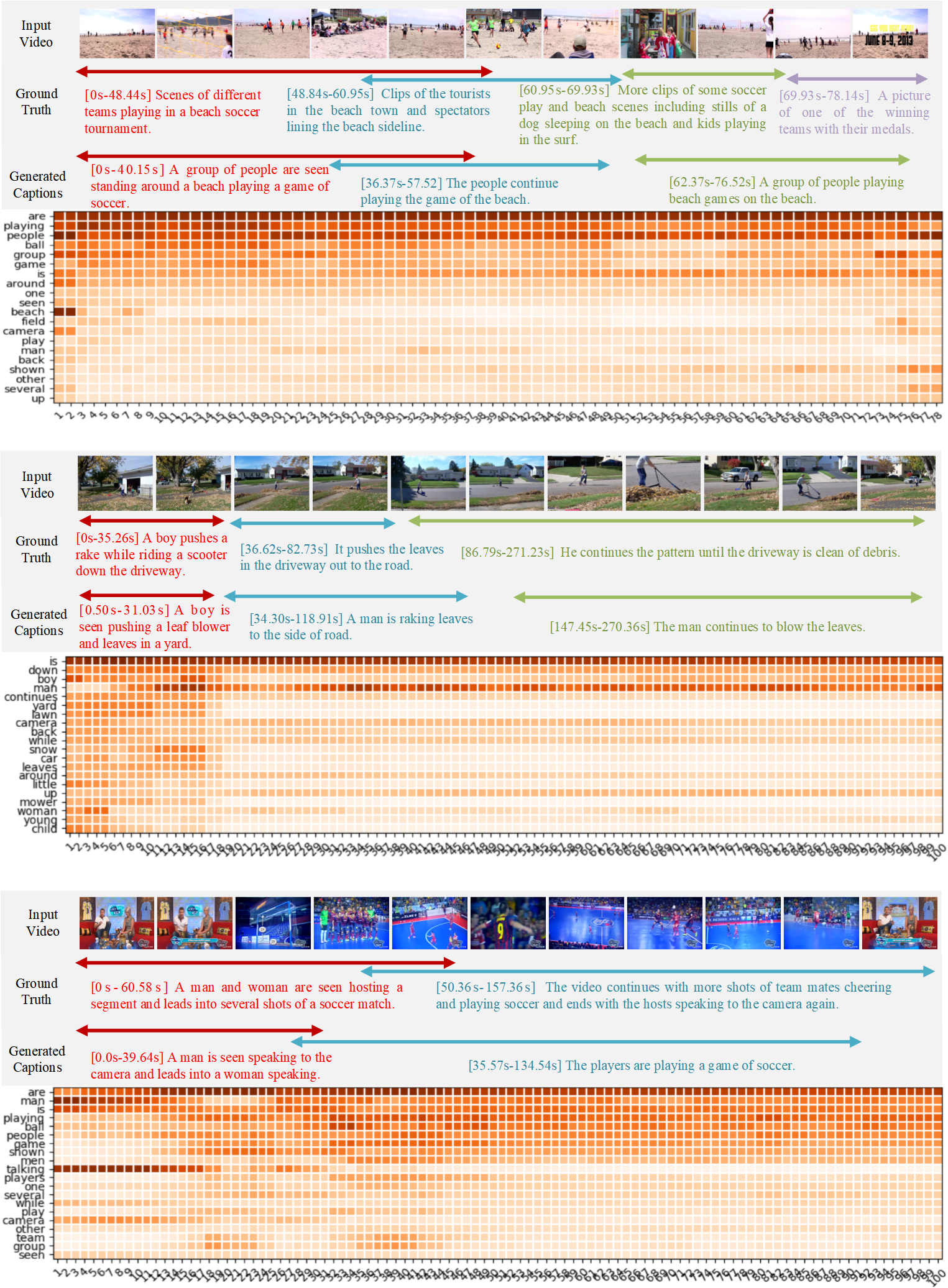}
	\caption{Examples of concept detection guiding video captioning. Note that for videos longer than 100 frames, we display concept probabilities only for the first 100 frames due to space limitations.}
	\label{fig:illustrate_concepts_supp}
\end{figure*}

As noted in the main paper, discerning relationships between objects solely through visual analysis remains challenging. Additionally, we observed that the construction of concept words influences caption quality, as irrelevant or redundant words can mislead the concept detection module and affect video captions. For instance, in the second video in Fig.\ref{fig:illustrate_concepts_supp}, the concept detection module predicted both `boy' and `man' for the same individual, resulting in inconsistencies across different event descriptions.

\end{document}